\documentclass{article}
\usepackage{odyssey_04,amsmath,amssymb,graphicx,url}
\ninept

\title{A comparison of linear and non-linear calibrations for speaker recognition}

\makeatletter
\def\name#1{\gdef\@name{#1\\}}
\makeatother
\name{{\em Niko Br\"ummer$^*$, Albert Swart$^*$ and David van Leeuwen$^+$}}
\address{$^*$AGNITIO Research, South Africa %{\small \tt nbrummer@agnitio-corp.com}  
\\$^+$Radboud University, Nijmegen, The Netherlands
}
\begin{document}
\maketitle
\begin{abstract}
In recent work on both generative and discriminative score to log-likelihood-ratio calibration, it was shown that linear transforms give good accuracy only for a limited range of operating points. Moreover, these methods required tailoring of the calibration training objective functions in order  to target the desired region of best accuracy. Here, we generalize the linear recipes to non-linear ones. We experiment with a non-linear, non-parametric, discriminative PAV solution, as well as parametric, generative, maximum-likelihood solutions that use Gaussian, Student's T and normal-inverse-Gaussian score distributions. Experiments on NIST SRE'12 scores suggest that the non-linear methods provide wider ranges of optimal accuracy and can be trained without having to resort to objective function tailoring.
\end{abstract}

\section{Introduction}
In our recent work on score calibration for speaker recognition, we employed \emph{linear} score-to-log-likelihood-ratio transforms, the parameters of which were trained via generative~\cite{DvL_IS13,NB_ICASSP14}, or discriminative~\cite{NB_IS13} methods. In both cases, we noticed that a linear transform could not calibrate well everywhere over a wide range of operating points. This meant we had to \emph{choose} in which operating region we wanted calibration to work best, by tailoring the training objective function. In both generative and discriminative cases, this was achieved (essentially) by artificially weighting the importance of target and non-target trials in the training data. In~\cite{DvL_IS13}, we used a weighted maximum-likelihood criterion, while in~\cite{NB_IS13}, a variety of different calibration-sensitive discriminative objective functions were explored.

While these strategies both resulted in good calibration in the targeted operating region, it also gave poorer calibration in other operating regions. In this paper, we explore the possibility of using more general, non-linear calibration transforms, with the hope that (i) they can give good calibration over a wider range of operating points and (ii) they can be trained without having to resort to specially tailored objective functions.

In what follows, we recapitulate our generative and discriminative linear calibration strategies, and then introduce several non-linear strategies. All of these are compared experimentally on scores from SRE'12. 

\section{Calibration}
We briefly summarize the calibration problem and some of its solutions. 

We consider a speaker recognizer that, when given speech input, outputs a raw score. The score should help to decide which of two hypotheses, known as \emph{target} and \emph{non-target} is true. The speech input has two parts, the enrollment speech and the test speech. The target hypothesis says the test speech is of the same speaker as the enrollment. The non-target hypothesis says the speakers are different.

In order to be able to use the recognizer to make cost-effective decisions, we can calibrate the recognizer scores, to give us \emph{log-likelihood-ratios} (LLRs)~\cite{art:brummer_csl_2006}. Calibration transforms a score, $s$, as:
\begin{align}
s\to\log\frac{P(s|H_1,B)}{P(s|H_2,B)}
\end{align}
where the likelihoods are conditioned on $H_1$, the target hypothesis, or $H_2$, the non-target hypothesis. The likelihoods are further conditioned on some background information, $B$, which may include generative or discriminative score modelling assumptions, model parameters, or data. If $B$ includes model parameters rather than data, the calibration method is known as a \emph{plug-in} method. If instead, $B$ contains data rather than parameters, the method is known as \emph{fully Bayesian}. In this paper, we shall work with plug-in methods, which perform well in situations where a large amount of training data is available, which is the case here. See~\cite{NB_ICASSP14} for an analysis of the relationship between plug-in and fully Bayesian solutions and~\cite{NB_Interspeech14} for an example where fully Bayesian calibration outperforms plug-in calibration when very little training is available. 

\subsection{Generative calibration}
\def\params{\lambda}
\def\Tset{\mathcal{T}}
\def\Nset{\mathcal{N}}

For training our parametric generative models, we shall use two flavours of maximum-likelihood (ML) criterion. Let $\Tset$ and $\Nset$ respectively denote the sets of target and non-target scores available for training. The \emph{plain} ML criterion is:
\begin{align}
\label{eq:plainML}
\sum_{s\in\Tset} \log P(s|H_1,\params) + \sum_{s\in\Nset} \log P(s|H_2,\params)
\end{align}
which is to be maximized w.r.t.\ $\params$, the calibration parameters. We use $\params$ to jointly denote the parameters of both target and non-target score distributions. In some cases, the parameters for the two distributions will be independent, so that $\params=(\params_1,\params_2)$, but in other cases, some of the parameters may be shared between the two distributions. 

The \emph{weighted} ML criterion is:
\begin{align}
\label{eq:gcrit}
\frac{\alpha}{T}\sum_{s\in\Tset} \log P(s|H_1,\params) + \frac{1-\alpha}{N}\sum_{s\in\Nset} \log P(s|H_2,\params)
\end{align}
where $T$ is number of targets, $N$ is number of non-targets, and $0<\alpha<1$ is a user-supplied relative weighting for targets vs non-targets. Notice that if $\alpha=\frac{T}{T+N}$, then the two criteria are equivalent. In~\cite{DvL_IS13}, we did linear calibration, which required weighted ML, which has the disadvantage that $\alpha$ has to be chosen appropriately. In this paper, we experiment with more general calibration models that can be trained with plain ML.

\subsection{Discriminative calibration}
For \emph{parametric} discriminative calibration, we choose a calibration function, denoted $\ell=f(s,\params)$, which maps scores, $s$, to LLRs, $\ell$. The parameters are trained by minimizing, w.r.t.\ $\params$, a criterion of the form:
\begin{align}
\label{eq:dcrit}
\frac{\alpha}{T}\sum_{s\in\Tset} C\left(f(s,\params),H_1\right) + \frac{1-\alpha}{N}\sum_{s\in\Nset} C\left(f(s,\params),H_2\right)
\end{align} 
where $C(\ell,H_i)$ is a special cost function known as a \emph{proper scoring rule}~\cite{NB_IS13}. Here $\alpha$ fulfils a similar function as in the weighted ML criterion. For the results reported here, we choose a linear calibration transform, so that $f(s,\params)=As+B$, where $\lambda=(A,B)$, while for $C$ we use the logarithmic proper scoring rule, which is equivalent to logistic regression---see~\cite{NB_IS13,art:Brummer_Fusion_TASLP_2007} for more details. 

For \emph{non-parametric} discriminative calibration, the calibration function $\ell=f(s)$ does not depend on a small number of parameters. Instead, it is allowed to vary freely within the class of all \emph{monotonic rising} functions from $\mathbb{R}$ to $\mathbb{R}$. It turns out that this class of functions is general enough that it does not matter any more which proper scoring rule is used, or what the value of $\alpha$ is. A calibration function can be found which simultaneously minimizes the discriminative criterion for all proper scoring rules and all values of $\alpha$, see~\cite{PhD,NBpav}. Moreover, an efficient algorithm, known as \emph{pool-adjacent-violators} (PAV),\footnote{PAV is also known as isotonic regression.} exists to find the calibration function~\cite{inproc:Zadrozny_PAV_2002}. 

\section{Evaluating goodness of calibration}
Our experimental setup is the same as in~\cite{NB_ICASSP14}. We performed all our calibration experiments on scores from a single speaker recognizer (an i-vector PLDA system), which was part of the ABC submission~\cite{ABC12} to the NIST SRE'12 speaker recognition evaluation~\cite{web:sre12}. 

We trained our calibration parameters on a large development set, having multiple microphone and telephone speech segments of male speakers from SRE'04, '05, '06, '08 and '10. This gave about 42 million scores, of which 0.07\% were targets, for calibration training. 

We tested the goodness of these calibrators on male speakers from the NIST SRE'12 extended trial set~\cite{web:sre12}, where we pooled all 5 common evaluation conditions, giving about 9 million trials, of which 0.1\% were targets.

For evaluation of goodness of calibration, we shall use \emph{normalized Bayes error-rate plots}~\cite{BOSARIS,PhD}. In these plots, we compute the error-rate that results when Bayes decisions are made by thresholding the to-be-evaluated LLR scores at the minimum-expected cost Bayes threshold. The $x$-axis represents the operating point in the form of \emph{prior log odds}, and the $y$-axis represents normalized Bayes-error rate:
\begin{align}
y&=\frac{pP_\text{miss}(\theta)+(1-p)P_\text{fa}(\theta)}{\min(p,1-p)}
\end{align}
where $p=\frac{1}{1+\exp(-x)}$ is a \emph{synthetic target prior}, while $P_\text{miss}(\theta)$ and $P_\text{fa}(\theta)$ are miss and false-alarm rates obtained when thresholding LLRs at the theoretically optimal \emph{Bayes threshold}, $\theta=-x$. The normalization factor, $\min(p,1-p)$ is the Bayes error-rate of a default recognizer that makes Bayes decisions based on the prior, $p$, alone. The $y$-axis can also be interpreted as the well-known \emph{detection cost function} (DCF) metric of the NIST Speaker Recognition Evaluation (SRE) series, provided the cost coefficients are set to unity. In addition to the \emph{actual} Bayes error-rate, $y$, every plot will also display the \emph{minimum} error-rate, $y'$, which can be obtained at the empirically optimal decision threshold at every operating point:
\begin{align}
y'&=\frac{\min_{\theta}pP_\text{miss}(\theta)+(1-p)P_\text{fa}(\theta)}{\min(p,1-p)} 
\end{align}
Again, $y'$ corresponds to min-DCF of the NIST SRE series. In all plots, $y'$ will be displayed as a dashed black line. 

To keep error-rates meaningful, we respect \emph{Doddinton's Rule of 30}~\cite{inproc:DoddingtonsRule_98}, by choosing the range of the $x$-axis so that there are always at least 30 errors of each kind (misses and false-alarms), at the empirically optimal threshold---see the discussions in~\cite{BOSARIS,PhD}.

\section{Linear calibrations}
\def\ND{\mathcal{N}}
We summarize and compare our previous linear calibration solutions of both generative and discriminative flavours. Both need prior-weighting to target the desired operating region.

\subsection{Generative: Gaussians with shared variance}
We repeat the calibration method of~\cite{DvL_IS13}. We let $\params=(\mu_1,\mu_2,v)$ and we assign Gaussian score distributions of the form:
\begin{align}
P(s|H_i,\params) = \ND(s|\mu_i,v)
\end{align}
where the $\mu_i$ are hypothesis-conditional means, but where the variance, $v$, is shared. This gives a linear calibration transform. To train $\params$, we use the weighted ML criterion~\eqref{eq:gcrit}. The maximizing parameters have simple, closed-form solutions---see~\cite{DvL_IS13} for details.

\begin{figure}[!htb]
\centerline{\includegraphics[width=0.5\textwidth,trim = 2cm 1.7cm 2cm 2cm,clip = true]{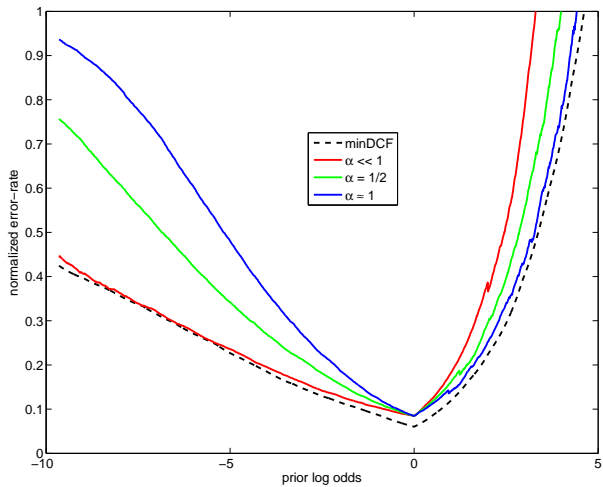}}
\caption{Accuracy of common-variance Gaussian calibration, using various values of ML weighting, $\alpha$.}  
\label{fig:cgauss}
\end{figure}

Figure~\ref{fig:cgauss} shows the calibration performance for the cases $\alpha=\frac{T}{T+N} = 0.0007$, $\alpha=\frac12$ and $\alpha = 0.92$. It is clear that good performance can be obtained locally by adjusting $\alpha$. Note the agreement between $\alpha$ (the training parameter) and $p=\frac{1}{1+e^{-x}}$ (the evaluation parameter). If $\alpha\ll1$, then performance is good for $p\ll1$, (on the negative $x$-axis). Conversely, if $\alpha\approx1$, then performance is good for $p\approx1$ (on the positive $x$-axis). Unfortunately, tuning for good performance in one place, causes higher error-rates elsewhere.

\subsection{Discriminative: logistic regression}
\begin{figure}[!htb]
\centerline{\includegraphics[width=0.5\textwidth,trim = 2cm 1.7cm 2cm 2cm,clip = true]{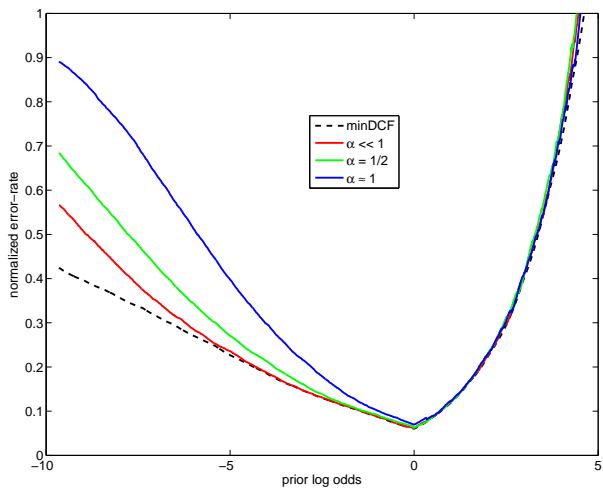}}
\caption{Accuracy of weighted logistic regression, with various values of weighting, $\alpha$.}  
\label{fig:lr}
\end{figure}

We repeat the linear logistic regression calibration of~\cite{NB_IS13}, trained with the weighted criterion~\eqref{eq:dcrit}. In figure~\ref{fig:lr} we show performance for weightings $\alpha=\frac{T}{T+N} = 0.0007$, $\alpha=\frac12$ and $\alpha = 0.92$. The same conclusions hold as in the generative case, except that performance near $x=0$ is better and is less sensitive to $\alpha$ on the positive $x$-axis.

The green plot in figure~\ref{fig:transfer} shows the linear logistic regression calibration transform for $\alpha=1/1000$.

\section{Non-linear calibrations}
Now we introduce the new work in this paper, namely several non-linear calibration strategies, none of which need objective function tailoring.

\subsection{Non-linear, discriminative: PAV}
\begin{figure}[!htb]
\centerline{\includegraphics[width=0.5\textwidth,trim = 2cm 1.7cm 2cm 2cm,clip = true]{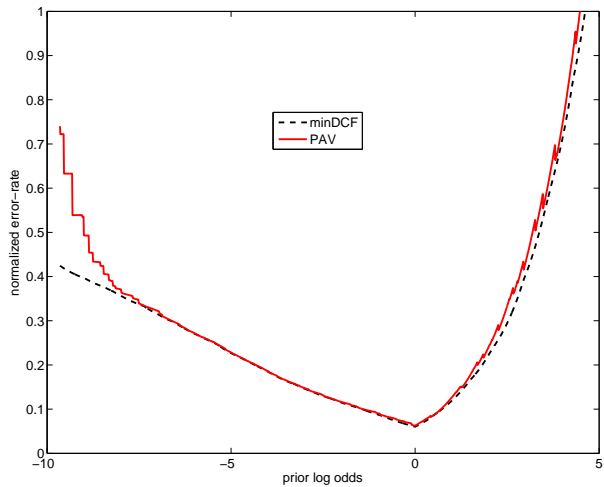}}
\caption{Accuracy of non-parametric, discriminative PAV calibration.}  
\label{fig:pav}
\end{figure}

The result of applying PAV calibration\footnote{A MATLAB implementation is available in the BOSARIS Toolkit at \url{sites.google.com/site/bosaristoolkit}.} is shown in figure~\ref{fig:pav}. There is only one solution, because there are no weighting parameters to tune. Very good calibration is obtained everywhere, except in the extreme left. The PAV calibration is optimal on the training data (for which indeed DCF equals min-DCF everywhere), but on the independent evaluation data (shown) calibration can be sub-optimal. We attribute the problem in the left to overtraining in this region, where there may not be enough false-alarms in the training data, relative to the rich choice of monotonic calibration functions. 

The red plot in figure~\ref{fig:transfer} shows the PAV calibration transform.

\subsection{Non-linear, generative: Gaussian distributions}
We obtain a non-linear (quadratic) calibration function by allowing separate variances for targets and non-targets, so that $\params=(\mu_1,v_1,\mu_2,v_2)$ and:
\begin{align}
P(s|H_i,\params) &= \ND(s|\mu_i,v_i)
\end{align}  
We now use the plain ML criterion~\eqref{eq:plainML}. The ML parameters have independent, closed-form solutions. The target parameters are:
\begin{align}
\mu_1 &= \frac{1}{T}\sum_{s\in\Tset} s, & 
v_1 &= \frac{1}{T} \sum_{s\in\Tset} (s-\mu_1)^2
\end{align}
and similar for non-targets. The accuracy is shown as the red plot in figure~\ref{fig:gen}. The blue plot in figure~\ref{fig:transfer} shows the (quadratic) Gaussian calibration transform.

\subsection{Non-linear, generative: T-distributions}
We generalize the Gaussian solution by adopting Student's T (or just T) distributions~\cite{book:Bishop_PRML}. Whereas the Gaussian distribution has two parameters (location and scale), the T distribution has three: location, scale and \emph{degrees of freedom} (d.o.f.). If the d.o.f.\ is large, the T distribution approaches the Gaussian. For small d.o.f., the distribution has heavy tails. 

Using independent distributions for targets and non-targets, the total number of parameters for this calibration model is 6 (3 each). Closed-form solutions for the ML parameters do not exist. One way to obtain an ML solution involves designing an EM-algorithm, based on a hidden scale variable associated with every score---see~\cite{KennyOdyssey10} for a similar EM-algorithm. However, we found our EM-algorithm was slow and prone to get stuck in saddle points, or other sub-optimal areas of small gradient. Instead, we found that direct, quasi-Newton numerical optimization, in our case BFGS~\cite{book:Nocedal_Optimization_2006}, was reliable and much faster. 

Since the basic BFGS algorithm that we used is an unconstrained optimizer\footnote{Modified versions of BFGS exist that can handle constraints.} and the scale and d.o.f.\ parameters are constrained to be positive, we needed to reparametrize those parameters via some suitable transform from $\mathbb{R}$ to the positive reals. We tested squaring and exponentiation. The former worked well, the latter not at all. 

The accuracy is shown in the green plot in figure~\ref{fig:gen}. The magenta plot in figure~\ref{fig:transfer} shows the T-distribution calibration transform.

\subsection{Non-linear, generative: NIG-distributions}
Finally, we generalize the Gaussian even futher, to a four-parameter family known as the \emph{normal-inverse-Gaussian} (NIG) distribution~\cite{NIG_Karlis}. The four parameters encode location, scale, skewness and tail weight. Using independent NIG parameters for targets and non-targets, the total number of parameters is 8. 

Although the ML solution may be sought via an EM-algorithm~\cite{NIG_Karlis}, we did not try it, preferring as before, direct optimization. In this case, however, we found that BFGS got stuck in saddle points. BFGS does not even know when it is in a saddle point, because it makes use of a positive-definite approximation to the Hessian.\footnote{In numerical optimization, it is customary to \emph{minimize} the objective function. In this case we minimize the negative likelihood. Minima have positive-definite Hessians (matrix of 2nd-order partial derivatives), but saddle points have Hessians with eigenvalues of mixed sign.} 

Next, we tried the more powerful \emph{trust-region-Newton} algorithm~\cite{book:Nocedal_Optimization_2006, conn2000trust}, which uses the true Hessian of the objective function. We computed the Hessian using the Pearlmutter trick~\cite{pearlmutter} and complex-step algorithmic differentiation~\cite{complexstep}. The first problem was that the Hessian computation took too long to perform over 42 million scores, because the NIG density requires the evaluation of Bessel functions~\cite{NIG_Karlis}. This was solved by using (for non-targets) small (1\%) randomly selected samples of the scores for the Hessian computation, but still using all the data for function value and gradient~\cite{stochastic_hessian}.

The second problem was that this algorithm still got stuck in saddle points. Simply trying to escape saddle points along the gradient did not help. What worked was to escape along the direction of the most negative curvature.\footnote{This is along the eigenvector corresponding to the most negative eigenvalue of the Hessian.}
 
The accuracy of the NIG solution is shown as the blue plot in figure~\ref{fig:gen}. Of the three generative non-linear calibration models, the NIG variant performs best on this data. We would however hesitate to recommend it before testing it on several other data sets. A disadvantage of the NIG solution is that it requires working with Bessel functions, which can be tricky and slow.

\begin{figure}[!htb]
\centerline{\includegraphics[width=0.5\textwidth,trim = 2cm 1.7cm 2cm 2cm,clip = true]{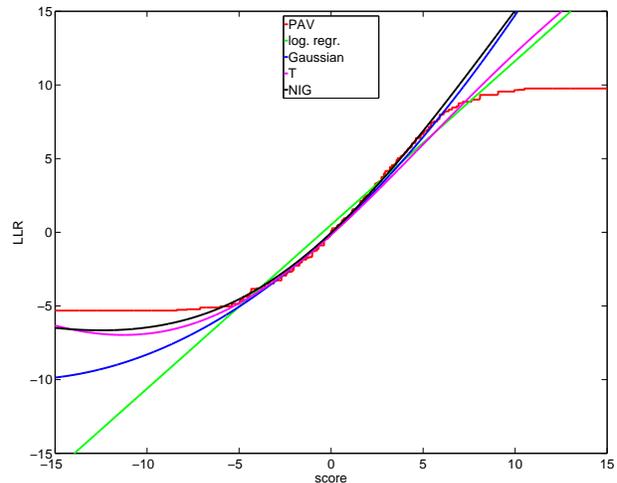}}
\caption{Comparison of score-to-LLR calibration transform functions.}  
\label{fig:transfer}
\end{figure}

\begin{figure}[!htb]
\centerline{\includegraphics[width=0.5\textwidth,trim = 2cm 1.7cm 2cm 2cm,clip = true]{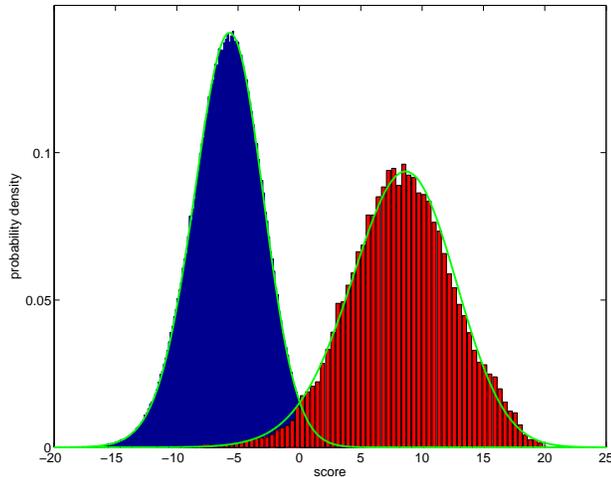}}
\caption{The maximum-likelihoood NIG solution, compared to normalized histograms of target and non-target scores.}  
\label{fig:hist}
\end{figure}

The black plot in figure~\ref{fig:transfer} shows the NIG calibration transform. Figure~\ref{fig:hist} shows the NIG probability densities (green) compared to histograms of the scores. 

\subsection{Discussion}
\begin{figure}[!htb]
\centerline{\includegraphics[width=0.5\textwidth,trim = 2cm 1.8cm 2cm 2cm,clip = true]{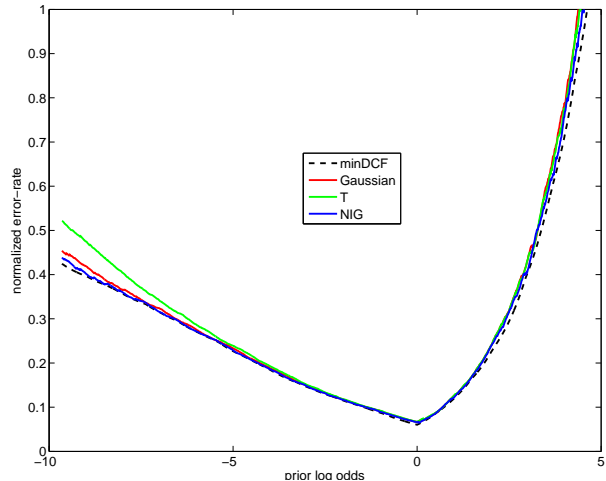}}
\caption{Accuracy of generative solutions: Gaussian, T and NIG distributions.}  
\label{fig:gen}
\end{figure}

Comparing Gaussian, T and NIG accuracies in figure~\ref{fig:gen}, the main differences are at operating points on the extreme left. It is perhaps surprising that the T-distribution (3 parameters), does worse than the Gaussian (2 parameters) and the NIG (4 parameters). One would expect that the more flexible T should be able to model the data more closely than the Gaussian, while being more immune to overtraining than the NIG. 

We speculate that this behaviour can be explained as follows. The Gaussian does not have the ability to accurately model the tails of the distributions and effectively ignores the tails. The T distribution has more flexible tail modelling capability, but being symmetric, it has to treat left and right tails the same. Effectively it will be compensating for skewness by making the tails thicker and this causes the observed inaccuracy. The NIG has an additional skewness parameter, so it does not have to use tail thickness to model skewness and can therefore model both tails more accurately.

It should also be mentioned  that this data suffers from mild dataset shift~\cite{book:datasetshift} (changes in score distributions between training and evaluation) and this complicates explanations of the observed accuracy.

\section{Conclusion}
We have shown that linear score-to-LLR calibration transformations struggle to give optimal accuracy over a wide range of operating points. If they are used, their training objective functions must be tailored to the desired operating region.

More flexible, non-linear calibrations can remain accurate over a wider range of operating points, while being trained with standard criteria that do not need to be tuned. 

The danger remains, as always, that more flexible recognizers can be more easily overtrained. In future work, we would like to investigate fully Bayesian methods as a safeguard against overtraining---see~\cite{NB_Interspeech14}, our first work in this direction, where we give a Bayesian solution for Gaussian score models. 

We would also like to explore the richer calibration models introduced here, for the problem of unsupervised calibration~\cite{NB_ICASSP14}.

\section{Acknowledgements}
We thank the anonymous reviewers for spotting the missing logarithms in~\eqref{eq:plainML} and~\eqref{eq:gcrit} and for asking a few interesting questions, some of which we were able to answer in the final version of this paper.

\bibliographystyle{IEEEbib}
\bibliography{refs}
%\begin{thebibliography}{10}
%\bibitem[1]{} tbd
%\end{thebibliography}
\end{document}